\documentclass[sn-vancouver]{sn-jnl}

\jyear{2021}%

\usepackage{algorithm}
\usepackage{latexsym}
\usepackage{amsmath}
\usepackage{colortbl}
\usepackage{tabularx,booktabs,multirow}
\usepackage{colortbl}
\usepackage{arydshln}
\usepackage{longtable}
\usepackage{ltablex}
\usepackage{subfig}

\definecolor{darkblue}{rgb}{0.0, 0.0, 0.55}
\definecolor{premise}{RGB}{40, 116, 166}
\definecolor{hypothesis}{RGB}{40, 116, 166}

\usepackage{bm}
\usepackage{pifont}
\newcommand{\xmark}{\ding{55}}%

\theoremstyle{thmstyleone}%
\theoremstyle{thmstyletwo}%

\theoremstyle{thmstylethree}%

\begin{document}

\title[LoNLI: Testing Diverse Logical Reasoning Capabilities for NLI]{\textsc{LoNLI}: An Extensible Framework for Testing Diverse \textsc{Lo}gical Reasoning Capabilities for \textsc{NLI}}


\author*[1]{\fnm{Ishan} \sur{Tarunesh}}\email{ishantarunesh@gmail.com}

\author[2]{\fnm{Somak} \sur{Aditya}}\email{saditya@cse.iitkgp.ac.in}

\author[3]{\fnm{Monojit} \sur{Choudhury}}\email{monojitc@microsoft.com}

\affil*[1]{\orgname{Samsung Electronics}, \city{Suwon}, \country{Korea}}

\affil[2]{\orgdiv{CSE}, \orgname{IIT Kharagpur}, \orgaddress{ \city{Kharagpur}, \country{India}}}

\affil[3]{\orgname{Microsoft Research}, \orgaddress{\street{9 Lavelle Road}, \city{Bangalore}, \country{India}}}


\abstract{Natural Language Inference (NLI) is considered a representative task to test natural language understanding (NLU). In this work, we propose an extensible framework to collectively yet categorically test diverse \textsc{Lo}gical reasoning capabilities required for \textsc{NLI} (and, by extension, NLU). Motivated by behavioral testing, we create a semi-synthetic large test bench ($363$ templates, $363k$ examples) and an associated framework that offers the following utilities: 1) individually test and analyze reasoning capabilities along 17 reasoning dimensions (including pragmatic reasoning); 2) design experiments to study cross-capability information content (leave one out or bring one in); and 3) the synthetic nature enables us to control for artifacts and biases. We extend a publicly available framework of automated test case instantiation from free-form natural language templates (CheckList) and a well-defined taxonomy of capabilities to cover a wide range of increasingly harder test cases while varying the complexity of natural language.
Through our analysis of state-of-the-art NLI systems, we observe that our benchmark is indeed hard (and non-trivial even with training on additional resources). Some capabilities stand out as harder. Further, fine-grained analysis and fine-tuning experiments reveal more insights about these capabilities and the models -- supporting and extending previous observations; thus showing the utility of the proposed testbench.}

\keywords{NLI, Reasoning, Benchmarking, Logic}



\maketitle

\section{Introduction}
Recently, the NLP community has observed that the measured performance improvements in the popular benchmarks do not always translate to targeted improvements in learning the capabilities expected for solving a task. This gap was specifically highlighted as \citet{probinglogical} and \citet{ribeiro-etal-2020-beyond} showed a collection of simple examples representing minimal expected capabilities for a task that is enough to degrade state-of-the-art models' performance completely. Such a \textit{performance degradation} gap was acknowledged and explicitly pointed out by \citet{DBLP:conf/naacl/BowmanD21}, who released a call for action for new benchmarks where the performance in benchmark aligns with good task performance. Furthermore, in \citet{Schlangen-2021}, authors point out crucial differences between benchmarking in computing and benchmarking as followed in NLP. Performing a task requires a collection of cognitive capabilities for a human. In contrast, datasets are only a reflection of the task, and models are often optimized on the train split of the benchmark dataset, irrespective of the required underlying cognitive capabilities. As highlighted by the authors, such a train-test setup makes it hard to know the connections between tasks and capabilities and the possible transfer of learned capabilities from one task to the other without re-training.

Several researchers believe a dynamic \textit{adversarially} created benchmark will accurately reflect performance, explaining the performance degradation. Nie et al. \citet{nie2019adversarial} proposed an adversarial human-in-the-loop dataset creation process where humans play a game with a state-of-the-art system(s) to create increasingly complex examples. This, however, does not sufficiently address the problem as the initial models still fail at simpler examples \citep{kaushik2019learning,probinglogical,ribeiro-etal-2020-beyond}, leaving the question ``\textit{can the system perform a particular type of reasoning?}'' unanswered. 

Inspired by behavioral testing, \cite{ribeiro-etal-2020-beyond} proposed the creation of template-based test-suites (called \textsc{CheckList}) that have broader coverage, ranging from minimal expected functionality to more complicated tests across a range of capabilities. Moving beyond phenomena-wise probing and cloze-task formulations, the methodology produces a capability-wise behavioral summary that aggregates different shortcomings of the SOTA models across capabilities in a disentangled manner. In this work, we show that with the right selection of required reasoning capabilities\footnote{We use the term ``capability'' following \textsc{CheckList}-introduced terminology \cite{ribeiro-etal-2020-beyond}. According to \cite{ribeiro-etal-2020-beyond}, a capability is simply a feature a model is expected to possess. Such capabilities may include logical reasoning abilities. Humans, on the other hand, \textit{may require} cognitive abilities to solve examples requiring such types of reasoning. However, the list of capabilities, as defined here, does not directly align with cognitive abilities.} such behavioral testing can help find out which reasoning abilities cause the performance degradation gap, thereby also establishing a task-capability relation \citep{Schlangen-2021}. A templated test suite following the behavioral testing methodology conforms to the traditional definition of \textit{benchmarking} in computing, where it is not necessary to provide a training split; rather might be discouraged, as behavioral functional testing is often independent of software development.

To this end, we propose a novel extensible Natural Language Inference benchmark to explicitly test a diverse set of reasoning capabilities. We create a \textsc{CheckList} that enables evaluation of whether NLI systems exhibit such reasoning capabilities. In the process, we extend the list of capabilities in \citet{ribeiro-etal-2020-beyond} to cover more interesting linguistic and logical reasoning phenomena (such as causal, spatial, pragmatic) required in NLI (and similar tasks such as Question-answering). We expand on how we come up with templates for such reasoning capabilities and discuss the different ways our benchmarking dataset can be used. Through a short survey of related datasets and our experiments, we show how the \textsc{LoNLI} dataset (and the associated framework) uniquely enables fine-grained disentangled capability-wise analysis along various dimensions; and, to some extent, address the \textit{performance degradation} by underscoring the harder capabilities. Identifying such \textit{harder} capabilities may further help develop targeted applications. For example, a system that does well on logical and numerical reasoning capability can perhaps be useful for designing a math solver (or tutor, such as the Microsoft Math App.), but not a general purpose reasoner which might require spatiotemporal reasoning and common sense reasoning.



\paragraph{Contributions.} Our contributions are as follows. We create a semi-synthetic capability-rich NLI test-suite \textsc{LoNLI} (17 capabilities, 363 templates, 363k examples) by extending the reasoning taxonomy of \citet{joshi2020taxinli}. We show that \textsc{LoNLI} is hard, extensible by nature, balanced across known dimensions of bias, and enables fine-grained analysis of reasoning abilities. From our analysis, we find 1) \textsc{LoNLI} is hard to solve even with additional targeted resources (such as \textsc{ImpPres} \citep{jeretic-etal-2020-natural}), 2) quantifier, implicature and \textit{deduction} capabilities (spatial, temporal, numerical) stand out as harder across systems; and 3) also orthogonal as they are hard to learn by sole training on other capabilities. 4) Further, template-wise and  intra-template observations attest to previously found facts and discover new ones without the hassle of targeted probing task creation.


\section{Survey of Existing Datasets}

The two most widely used large-scale NLI datasets are the SNLI \citep{bowman2015large} and the MultiNLI (MNLI) dataset \citep{bowman2018-mnli}. Researchers have found that the high performance of the Transformer-based pre-trained language models (PTLM) on these datasets (and corresponding NLU benchmarks which include them) is often due to annotation biases and spurious artifacts in the datasets \citep{Poliak2018HypothesisOB, Gururangan2018AnnotationAI}. Such findings and performance degradation of trained PTLMs on targeted phenomena-wise tests have led to different evaluation strategies, such as i) probing tasks, ii) challenge datasets, and iii) capability-wise fine-grained evaluation. Effectiveness of probing tasks \citep{tenney2019bert,tenney2019you,hewitt-manning-2019-structural,jawahar-etal-2019-bert,liu2019linguistic,kim-etal-2019-probing} and challenge datasets \citep{mccoy-etal-2019-right,kaushik2019learning} are limited when it comes to answering the problem of performance degradation on simple examples or finding out comparatively harder capabilities using a single benchmark. 
Acknowledging this, researchers have built datasets focusing on testing a (limited) range of reasoning capabilities \citep{poliak-etal-2018-collecting,probinglogical}, which include examples of varying complexity. \cite{ribeiro-etal-2020-beyond} has taken an important step towards formalizing the notion of testing, and proposed the \textsc{CheckList} framework for NLP tasks. Their results on several tasks have made the above performance degradation explicit, as PTLMs were shown to fail at minimal expected capabilities.

\begin{table}
\newcolumntype{L}[1]{>{\raggedright\let\newline\\\arraybackslash\hspace{0pt}}m{#1}}
\newcolumntype{C}[1]{>{\centering\let\newline\\\arraybackslash\hspace{0pt}}m{#1}}
\setlength\fboxsep{1pt}
\resizebox{\textwidth}{!}{%
\scriptsize
\setlength\tabcolsep{2pt}
\begin{tabular}{L{0.15\columnwidth} L{0.25\columnwidth} C{0.1\columnwidth} C{0.5\columnwidth}}
\toprule
\multirow{2}{*}{Dataset} & Creation & \multirow{2}{*}{Size} & \multirow{2}{*}{Reasoning Types}\\
& Methodology & & \\
\midrule
SNLI \tiny\cite{bowman2015large} & Crowd-sourced & 570K & \xmark \\
\arrayrulecolor{black!20}\hdashline\addlinespace[3pt]
\multicolumn{4}{L{\columnwidth}}{P: A black race car starts up in front of a crowd of people. H: A man is driving down a lonely road. {\texttt{(neutral)}}} \\
\midrule
MNLI \tiny\cite{bowman2018-mnli} & Crowd-sourced & 433K & \xmark \\
\arrayrulecolor{black!20}\hdashline\addlinespace[3pt]
\multicolumn{4}{L{\columnwidth}}{P: At the other end of Pennsylvania Avenue, people began to line up for a White House tour. H: People formed a line at the end of Pennsylvania Avenue. {\texttt{(entail)}}}\\
\midrule
QNLI \tiny\cite{rajpurkar-etal-2016-squad} & Crowd-sourced & 110k & \xmark \\
\arrayrulecolor{black!20}\hdashline\addlinespace[3pt]
\multicolumn{4}{L{\columnwidth}}{P: What library was estimated to have 700,000 volumes? H: The city of Pergamon also had a large library and became a major center of book production. {\texttt{(not\_entail)}}}\\
\midrule
ANLI \tiny\cite{nie2019adversarial} & \begin{tabular}[c]{@{}l@{}}Crowd-sourced with \\ human in the loop\end{tabular} & 162K & \xmark \\
\arrayrulecolor{black!20}\hdashline\addlinespace[3pt]
\multicolumn{4}{L{\columnwidth}}{P: Javier Torres (born May 14, 1988, in Artesia, California) is an undefeated Mexican American professional boxer in the Heavyweight division. Torres was the second-rated U.S. amateur boxer in the Super Heavyweight division and a member of the Mexican Olympic team. H: Javier was born in Mexico {\texttt{(contradict)}}}\\
\midrule
WNLI \tiny\cite{levesque2012winograd} & Expert-annotated & 780 & Coreference \\
\arrayrulecolor{black!20}\hdashline\addlinespace[3pt]
\multicolumn{4}{L{\columnwidth}}{P: The cookstove was warming the kitchen, and the lamplight made it seem even warmer. H: The lamplight made the cookstove seem even warmer. {\texttt{(Coreference, contradict)}}}\\
\midrule
ConTRoL \tiny\cite{Liu2021NaturalLI} & Expert annotated & 8325 & Coreference, Logical, Temporal, Analytical, Information Integration\\
\arrayrulecolor{black!20}\hdashline\addlinespace[3pt]
\multicolumn{4}{L{\columnwidth}}{P: The biggest risk facing the world's insurance companies is possibly the rapid change now taking place within their own ranks. Sluggish growth in core markets and intense price competition, coupled with shifting patterns of customer demand and the rising cost of losses, are threatening to overwhelm those too slow to react. H: Insurance companies are experiencing a boom in their core markets. {\texttt{(Coreference, contradict)}}}\\
\midrule
FaNCY \tiny\cite{DBLP:conf/clic-it/RocchiettiAMSL21} & Manual & 4k & Factivity, Negation, Commonsense, Taxonomic\\
\arrayrulecolor{black!20}\hdashline\addlinespace[3pt]
\multicolumn{4}{L{\columnwidth}}{P: All seagulls fly. H: All birds fly. {\texttt{(Taxonomic, neutral)}}}\\
\arrayrulecolor{black!20}\hdashline\addlinespace[3pt]
\multicolumn{4}{L{\columnwidth}}{P: The man was born in 1950. H: The man was 18 in 1968. {\texttt{(Commonsense, entail)}}}\\
\midrule
CLCD \tiny\cite{clcd2019} & Synthetic & 60K & Logical \\
\arrayrulecolor{black!20}\hdashline\addlinespace[3pt]
\multicolumn{4}{L{\columnwidth}}{P: Ashley is as tall as Sonya, Sonya is as tall as Carol, Carol is as tall as Miriam, Miriam is as tall as Jan, Jan is as tall as Loretta, Loretta is taller than Edith H: Carol is taller than Edith {\texttt{(Comparative, entail)}}}\\
\arrayrulecolor{black!20}\hdashline\addlinespace[3pt]
\multicolumn{4}{L{\columnwidth}}{P: Lula has visited Valdez, Lauren has visited Sunnyvale, Vickie has visited Ponca City H: 	Lula didn't visit Yuma {\texttt{(Negation, entail)}}}\\
\midrule
\textsc{ImpPres} \tiny\cite{jeretic-etal-2020-natural} & Synthetic & 25K & Pragmatic \\
\arrayrulecolor{black!20}\hdashline\addlinespace[3pt]
\multicolumn{4}{L{\columnwidth}}{P: All teenagers criticize Guy. H: Some teenagers criticize Guy. {\texttt{(Implicature (some - all), contradict)}}}\\
\midrule
\textsc{HANS} \tiny\cite{mccoy-etal-2019-right} & Synthetic & 30K & Lexical, Syntactic \\
\arrayrulecolor{black!20}\hdashline\addlinespace[3pt]
\multicolumn{4}{L{\columnwidth}}{P: The president advised the doctor. H: The doctor advised the president. {\texttt{(Lexical (sub-obj swap), not\_entail)}}}\\
\midrule
\end{tabular}
}
\end{table}

\begin{table}[!ht]
\newcolumntype{L}[1]{>{\raggedright\let\newline\\\arraybackslash\hspace{0pt}}m{#1}}
\newcolumntype{C}[1]{>{\centering\let\newline\\\arraybackslash\hspace{0pt}}m{#1}}
\setlength\fboxsep{1pt}
\resizebox{\textwidth}{!}{%
\scriptsize
\setlength\tabcolsep{2pt}
\begin{tabular}{L{0.15\columnwidth} L{0.25\columnwidth} C{0.1\columnwidth} C{0.5\columnwidth}}
\toprule
Vashishtha et al. \tiny\cite{vashishtha-etal-2020-temporal} & Synthetic & 1.1M & Temporal \\
\arrayrulecolor{black!20}\hdashline\addlinespace[3pt]
\multicolumn{4}{L{\columnwidth}}{P:  The greeter said there was about 15 mins waiting. H: The saying did take or will take shorter than an hour. {\texttt{(Temporal (duration), entail)}}}\\
\midrule
TaxiNLI \tiny\cite{joshi2020taxinli} & \begin{tabular}[c]{@{}l@{}} Crowd-annotated \\(Subset of MNLI)\end{tabular} & 10K & Lexical, Syntactic, \textit{Connective}, \textit{Deductions}, \textit{Knowledge} \\
\arrayrulecolor{black!20}\hdashline\addlinespace[3pt]
\multicolumn{4}{L{\columnwidth}}{P: Some travelers add Molokai and Lanai to their itineraries. H: No one decides to go to Molokai and Lanai. {\texttt{(Quantifier, contradict)}}}\\
\arrayrulecolor{black!20}\hdashline\addlinespace[3pt]
\multicolumn{4}{L{\columnwidth}}{P: Actually, my sister wrote a story on it. H: My sibling created a story about it. {\texttt{(Relational, entail)}}}\\
\arrayrulecolor{black!20}\hdashline\addlinespace[3pt]
\multicolumn{4}{L{\columnwidth}}{P: See you Aug. 12, or soon thereafter, we hope. H: The person told not to come until December. {\texttt{(Temporal, entail)}}}\\
\midrule
\textbf{\textsc{LoNLI} (ours)} & Semi-Synthetic & 363K & Lexical, Syntactic, \textit{Connectives}, \textit{Deductions}, \textit{Numeric}, Pragmatic, \textit{Knowledge}\\
\arrayrulecolor{black!20}\hdashline\addlinespace[3pt]
\multicolumn{4}{L{\columnwidth}}{P: Robin and Mary are from Ukraine and Chile respectively. H: Robin is from Chile. {\texttt{(Boolean, contradict)}}}\\
\arrayrulecolor{black!20}\hdashline\addlinespace[3pt]
\multicolumn{4}{L{\columnwidth}}{P: Florence has six dollars. She gave away four dollars. H: Florence now has two dollars. {\texttt{(Numerical, entail)}}}\\
\arrayrulecolor{black!20}\hdashline\addlinespace[3pt]
\multicolumn{4}{L{\columnwidth}}{P: Irving is 350 miles from San Francisco and 570 miles from Miami. H: Irving is nearer to San Francisco than Irving is to Miami. {\texttt{(Spatial, entail)}}}\\
\arrayrulecolor{black!20}\hdashline\addlinespace[3pt]
\multicolumn{4}{L{\columnwidth}}{P: Marilyn's daughter is meticulous. H: Marilyn has a daughter. {\texttt{(Presupposition, entail)}}}\\
\arrayrulecolor{black}\bottomrule
\end{tabular}
}
\caption{The position of \textsc{LoNLI} in the spectrum of (a subset of closely related) NLI datasets. The \textit{Connective}, \textit{Deductions}, and \textit{Knowledge} are the buckets of reasoning types according to \textsc{TaxiNLI} taxonomy.}
\label{tab:survey}
\end{table}

Based on such observations, many calls \citep{Schlangen-2021,DBLP:conf/naacl/BowmanD21} for proposing improved NLP benchmarks have poured in. The following properties for an ideal NLU benchmark seems essential: 1) Benchmarks should make the connection between task and required capabilities explicit, where capabilities should be transferable across tasks (NLI, QA). 2) Benchmarks should be hard to solve, and 3) extensible. 4) Benchmarks should control for annotation artifacts and known biases. 5) Benchmarks should enable an understanding of the limits of current systems as well as whether systems exhibit minimal expected functionalities. Next, we will discuss some of the relevant existing NLI benchmark datasets (with and without reasoning type annotations). We observe that most recent datasets fall short on one or multiple of these criterion, delineating the necessity of our testbench.

 \textbf{Capability-agnostic} datasets such as QNLI, SciTail \citep{DBLP:conf/aaai/KhotSC18}, and FEVER \citep{DBLP:conf/naacl/ThorneVCM18} focus on a specific aspect of knowledge (such as knowledge of scientific phenomenon or world knowledge). Datasets such as FEVER often exhibit biases \citep{schuster-etal-2019-towards}, possibly inherited from the source. Moreover, most examples in these datasets (even targeted ones such as WNLI)  often require multiple capabilities.  DialogueNLI \citep{welleck-etal-2019-dialogue} features a persona-based dialogue structure for making inferences on the current utterance based on previous dialogue history. 
ANLI \citep{nie2019adversarial} is created through a human-in-the-loop process where participants play an adversarial game to fool a state-of-the-art system. While the methodology yields a way to create dynamic and increasingly harder benchmarks, the resulting dataset becomes model-aware and does not help address the performance degradation on simpler examples. Examples in ANLI also lack explicit capability information. 
\textbf{Capability-aware} datasets predominantly fall under two sub-types. The first type includes datasets created manually (FaNCY \citep{DBLP:conf/clic-it/RocchiettiAMSL21}) or requires experts for annotations (WNLI \citep{levesque2012winograd}, ConTRoL \citep{Liu2021NaturalLI}) with the exception of TaxiNLI \citep{joshi2020taxinli} which was annotated using crowdsourced workers. The second type includes datasets which are synthetic (CLCD  \citep{salvatore2019logical}, \textsc{ImpPres} \citep{jeretic-etal-2020-natural})  or semi-synthetic (\textsc{LoNLI} (ours)) in nature. Examples in ConTRoL test understanding of coreference resolution, logical and temporal reasoning along with information integration over multiple sentences. The examples are  expert-designed using publicly available online practice tests as the source. 
In TaxiNLI, authors annotate a subset of MNLI with taxonomic information rather than creating a synthetic dataset. However, authors do not control for the biases and artifacts that may have been inherited from the MNLI, and it is also hard to extend as annotation require extensive training. 
A wide array of past works use synthetically generated datasets for the purpose of capability-wise testing. 
\cite{probinglogical} create a probing dataset to test logical reasoning. 
\cite{mccoy-etal-2019-right} propose the HANS dataset for testing lexical and syntactic phenomena (subsequence and constituent). \cite{vashishtha-etal-2020-temporal} created a dataset aimed at testing temporal reasoning.  CLCD \citep{clcd2019} covers logical reasoning capabilities such as boolean, quantifiers, connectives, and counting. The \textsc{ImpPres} \citep{jeretic-etal-2020-natural} dataset is aimed at testing pragmatic capabilities such as implicature and presupposition. The recent work of \cite{yang-etal-2022-testaug} provides a two-step framework (TestAug) for generating test examples using templates assisted by GPT-3.
Other datasets include AlphaNLI \citep{Bhagavatula2020Abductive} (cloze-style dataset rather than NLI) explores the problem of abductive reasoning - finding the most plausible narrative context given the outcome, 

In contrast to existing datasets (Tab.~\ref{tab:survey}), through the proposed \textsc{LoNLI} benchmark, we enable evaluation at multiple granularity levels: intra-template (grouping via lexicon), a template, a (reasoning) capability, or the entire test bench level. We extend the capabilities proposed in \cite{joshi2020taxinli}, covering a gamut of reasoning abilities required for the NLI task (and NLU in general). Unlike challenge datasets, our dataset is not hard by design as the templates are targeted to cover minimal functionalities. But upon evaluation, we observe that our benchmark is difficult for most SOTA systems (and non-trivial to solve even with additional resources). Despite having simple templates we are able to attest existing observations from past works and also find out new observations. This brings us to another merit, i.e., extensibility. \textsc{LoNLI} can be extended in multiple dimensions: (i) more capability and templates can be added for probing particular phenomena, and (ii) templates involving multiple capabilities (hardness). Harder templates, in a cognitive sense, can be created by proposing templates that require multiple capabilities. 
Lastly, being semi-synthetic in nature, LoNLI is balanced with respect to known dimensions of  biases, such as \textit{gender-profession} bias \citep{Manela2021StereotypeAS},  and lexical bias \citep{Gururangan2018AnnotationAI} such as ``not'' being strongly correlated with contradiction (see Section \ref{sec:observations}). 

\section{A \textsc{Lo}gical \textsc{CheckList} for \textsc{NLI}}
The \textsc{CheckList}\footnote{\url{https://github.com/marcotcr/checklist}} methodology \citep{ribeiro-etal-2020-beyond} assists users in testing NLP models by creating templates for a variety of linguistic capabilities. These templates can then be used to generate multiple examples following the same structure but with different values of placeholders.
Table~\ref{tab:template1} shows an example \textsc{CheckList} template for NLI task.
Here \textcolor{darkblue}{NAME1} and \textcolor{darkblue}{NAME2} are placeholders of the same lexicon \textcolor{darkblue}{\{NAME\}} = \{Alexia, John, Mia, ...\} and \textcolor{darkblue}{ADJ} comes from the lexicon \textcolor{darkblue}{\{ADJ\}} = \{good, bad, kind, ...\}.
\begin{table}[!ht]
\resizebox{\columnwidth}{!}{%
\begin{tabular}{@{}lc@{}}
\toprule
\multicolumn{1}{c}{Template} & Expected \\ \midrule
P: \textbf{\big\{NAME1\big\}} and \textbf{\big\{NAME2\big\}} are \textbf{\big\{ADJ\big\}}. H: \textbf{\big\{NAME1\big\}} is \textbf{\big\{ADJ\big\}}. & Entail \\
Example: P: \underline{John} and \underline{Harry} are \underline{apathetic}. H: \underline{John} is \underline{apathetic}. & Entail \\ 
\bottomrule
\end{tabular}%
}
\caption{An example NLI template testing connective \textbf{and}.}
\label{tab:template1}
\end{table}
In tasks such as NLI, inferencing often requires (one or more) linguistic and logical reasoning capabilities. Our goal is to test the systems against such reasoning types.
This presents us with two challenges for templated
test-suite generation for the NLI task (or tasks that require different types of reasoning): 1) careful selection of capabilities that represent well-known linguistic and logical taxonomy, and are easily extensible, 2) template creation for such capabilities. 


\subsection{Selection of Capabilities}
\label{subsec:taxinli}

Our central aim is to perform capability-wise analysis of model performance. Hence, we selected a list of task-agnostic capabilities, 1) that have minimal overlap and can therefore aid analysis, 2) easily extensible, 3) relevant and covers a wide range of existing phenomena in public NLI datasets. Available categorizations in the literature are often dataset and task-specific, and many focus on system-specific error analysis.  GLUE Diagnostic test and the Adversarial NLI categories \citep{wang2018glue,nie2019adversarial} are good examples of error categorization. Comparably generic categorizations such as InfoTabs \citep{gupta-etal-2020-infotabs}, and \textsc{CheckList} \citep{ribeiro-etal-2020-beyond}, often conflate different granularities for categorizations; based on the present aim of the paper or the task(s) under consideration. For example, InfoTabs consider ``Named Entities'' as a category as well as ``Temporal'', and ``Negation''. Based on the definition by authors, it is not clear which high-level linguistic or logical category, ``Named Entities'' may belong to. Most of these categorizations are also highly overlapping, and therefore extending with more capabilities and templates become harder. Rather, we look towards first principles and start with the high-level categories as informed by Linguistics \citep{wittgenstein-1922,Jurafsky+Martin:2009a} and Logic \citep{sowa2010role}. Many Linguistics researchers argue that  deciphering \textit{meaning} from natural language \textit{form} often takes Lexical, Syntactic, Semantic, and Pragmatic abilities. From the perspective of Logic (Charles Pierce), there are three pre-dominant forms of reasoning: deductive, inductive, and abductive; that can be employed by an agent to \textit{understand} language. We observe that authors in \cite{joshi2020taxinli} proposed a high level categorization that attempts to combine both categorizations, and defined an extensible Taxonomy (\textsc{TaxiNLI}) suitable for our purposes.


\paragraph{\textsc{TaxiNLI}.}
\cite{joshi2020taxinli} proposed an (extensible) categorization of the reasoning tasks involved in NLI with a granularity finer than the high-level categorizations (Language and Logic), based on their relevance with respect to current public NLI datasets. Authors define three broad groups of reasoning: {\sc Linguistic}, {\sc Logical} and {\sc Knowledge}, which aligns with our philosophy.  The {\sc Linguistic} category covers examples where inference process is internal to the provided text; capturing lexical and syntactic aspects of meaning. As \textit{Semantic} category is broad and may require various forms of reasoning, authors define the high-level categories as {\sc Logical} and {\sc Knowledge}. {\sc Logical} is defined as the set of examples, where the inference process is external to the text, but may not require external facts to reason with. The {\sc Knowledge} category subsumes examples that require reasoning with facts and knowledge external to the input text. The examples under the {\sc Logical} category considers deductive reasoning mostly, and some examples under the {\sc Knowledge} category considers abductive reasoning. By definition, the categories in \textsc{TaxiNLI} have minimal overlap, and are extensible as they are arranged in a well-defined hierarchical Taxonomy. Lastly, authors have retained (or selected) a lowest-level category only when manual analysis demonstrated that a significant number of examples in public NLI datasets require that form of reasoning. Hence, the lowest-level categories constitute a suitable starting point.


 In Tab.~\ref{tab:exampletemplates}, we further introduce the categories and corresponding examples from the taxonomy in \cite{joshi2020taxinli}\footnote{For detailed definitions, please see \cite{joshi2020taxinli}.}.
{\sc Linguistic} category is further divided into capabilities \underline{lexical} and \underline{syntactic}. Authors have minimized overlap by carefully defining each category. For example, \underline{lexical} reasoning denotes examples where the Premise (P)-Hypothesis (H) pairs only differ through addition, removal or substitution of a few lexical items. \underline{Syntactic} examples are where understanding of syntactic variations or paraphrases is necessary for the inference process. 
{\sc Logical} categories are grouped under {\it Connectives},  {\it Mathematical} and {\it Deduction}. {\it Connectives} involve categories such as \underline{negation}, \underline{boolean}, \underline{quantifier}, \underline{conditional} and \underline{comparative}. They each target a specific logical operator (negation, and/or), if-else constructs (conditional), or the quantification phenomena (such as some, all, every). {\it Mathematical} includes examples requiring reasoning about quantities such as \underline{counting}, \underline{arithmetic}. \cite{joshi2020taxinli} had removed it for \textsc{TaxiNLI} annotation, as pilot studies showed the number of examples requiring mathematical reasoning in MultiNLI is limited.  \textit{Deduction} comprises of reasoning pertaining to \underline{relational}, \underline{spatial}, \underline{temporal}, \underline{causal} and \underline{coreference}. Relational reasoning requires understanding of relationships between entities mentioned in text. Spatial and temporal pertains to reasoning about time and space. Causal focuses on examples involving the understanding of cause-effect, while coreference involves examples requiring anaphora resolution. {\sc Knowledge} indicates examples where external (\underline{world}) or commonly assumed (we only consider \underline{taxonomic}) knowledge is required for inference.  We extend the taxonomy by adding back the (pruned) \underline{numerical} category. We also add a high-level category {\sc Pragmatic} (at the same level as {\sc Linguistic}, {\sc Logical}, and {\sc Knowledge}) to capture pragmatic phenomena. There are various types of pragmatic phenomena according to Grice's cooperative principle and other linguistic theories for interpretaion of utterances. Here, we include two basic sub-categories \underline{pre-supposition} (subsuming \underline{factivity}) and \underline{scalar implicature}, which can be further added to in future.

\begin{table*}[!ht]
\setlength\fboxsep{1pt}
\setlength\aboverulesep{0pt}
\setlength\belowrulesep{0pt}
\resizebox{\textwidth}{!}{%
\scriptsize
\setlength\tabcolsep{2pt}
\begin{tabular}{p{0.70cm} p{14.5cm} c}
\toprule
\# & \multicolumn{1}{c}{Description of phenomenon \& Examples} & Annotations\\
\arrayrulecolor{black}\midrule

\multirow{3}{*}{T1} & Ordered resolution of length 2 with matched hypothesis (Boolean) &\multirow{3}{*}{E (1)}\\
& P: George and Michael are from Germany and Australia respectively. H: George is from Germany. {\texttt{entail}}\\
\arrayrulecolor{black}\midrule
\multirow{2}{*}{T2} & Ordered resolution of length 2 with mis-matched hypothesis (Boolean) &\multirow{2}{*}{C (1)}\\
& P: Helen and Barbara are from Canada and Brazil respectively. H: Helen is from Brazil. {\texttt{contradict}}\\
\arrayrulecolor{black}\midrule
\multirow{3}{*}{T3} & Coreference resolution between gendered names (Coreference) &\multirow{3}{*}{E (.8)}\\
& P: Ricardo and Angelique are collegues. He is a minister and she is a model. H: Angelique is a model. {\texttt{entail}}\\
\arrayrulecolor{black}\midrule
\multirow{3}{*}{T4} & Comparison of distance between locations and infering near/far (Spatial) &\multirow{3}{*}{E (1)}\\
& P: Manchester is 67 miles from Pittsburg and 27 miles from Kansas. H: Manchester is nearer to Kansas than Manchester is to Pittsburg. {\texttt{entail}}\\
\arrayrulecolor{black}\midrule
\multirow{3}{*}{T5} & Infering the first or last event given a sequence of events (Temporal) &\multirow{3}{*}{E (1)}\\
& P: Jessica has latin class then geography class then french class. H: Latin class is the first class among the three. {\texttt{entail}}\\
\arrayrulecolor{black}\midrule
\multirow{2}{*}{T6} & Understanding of casual verb pair (such as give-take, throw-catch) (Causal) &\multirow{2}{*}{E (1)}\\
& P: Fiona taught mathematics to Ruth. H: Ruth learnt mathematics from Fiona. {\texttt{entail}}\\
\arrayrulecolor{black}\midrule
\multirow{2}{*}{T7} & Basic counting (Numerical) &\multirow{2}{*}{E (1)}\\
& P: Dorothy, Katie and Rose are the only children of Stephen. H: Stephen has exactly 3 children. {\texttt{entail}}\\
\arrayrulecolor{black}\midrule
\multirow{2}{*}{T8} & Basic counting (Numerical) &\multirow{2}{*}{C (1)}\\
& P: Martin is the child of Susan. Jill is the child of Susan. H: Susan has atleast 2 children. {\texttt{contradict}}\\
\arrayrulecolor{black}\midrule
\multirow{2}{*}{T9} & Knowledge of city and the countries they belong to (World) &\multirow{2}{*}{E (.8)}\\
& P: Patrick lives in Lahore. H: Patrick lives in Pakistan {\texttt{entail}}\\
\arrayrulecolor{black}\midrule
\multirow{2}{*}{T10} & Existence of hypothesis based on presupposition trigger in premise. (Pre-Supposition)&\multirow{2}{*}{E (1)} \\
& P: William's wife is tactful. H: William has a wife. {\texttt{entail}} \\
\arrayrulecolor{black}\midrule
\multirow{2}{*}{T11} & Existence of hypothesis based on presupposition trigger in premise.  (Pre-Supposition) &\multirow{2}{*}{E (1)}\\
& P: Martin has stopped drinking. H: Martin used to drink. {\texttt{entail}}\\
\arrayrulecolor{black}\midrule
\multirow{3}{*}{T12} & Change of quantifier from positive (some) to negative (not all).  (Implicature) &\multirow{3}{*}{E (1)} \\
& P: Some of the banknotes are crimson in colour. H: Not all of the banknotes are crimson in colour. {\texttt{entail}}\\
\arrayrulecolor{black}\midrule
\multirow{3}{*}{T13} & Testing maximum informativeness in utterance (Implicature) &\multirow{2}{*}{N (.5) E (.5)}\\
& P: Toothpaste and eyeliner lie on the table. Jane asked for eyeliner. H: Jane did not ask for the toothpaste. {\texttt{entail}}\\
\arrayrulecolor{black}\midrule
\multirow{2}{*}{T14} & Testing maximum informativeness in utterance (Implicature) &\multirow{2}{*}{E (1)}\\
& P: Donald asked Chris for 200 dollars. Chris had 90 dollars. H: Chris didn't have 200 dollars. {\texttt{entail}}\\
\arrayrulecolor{black}\midrule
\multirow{2}{*}{T15} & Testing ellipsis (syntactic), antonym (lexical) and comparative (Multiple) & \multirow{2}{*}{E (1)} \\ 
& P: Evelyn is weaker than Billy but not Michael. H: Evelyn is stronger than Michael. {\texttt{entail}}\\
\arrayrulecolor{black}\midrule
\multirow{3}{*}{T16} & Testing numerical comparison and commonsense (Multiple) &\multirow{3}{*}{C (1)}\\
& P: Grace asked Albert for 20 dollars. Albert refused. H: Grace has higher chances of convincing Albert for 40 dollars. {\texttt{contradict}} \\
\arrayrulecolor{black}\bottomrule
\end{tabular}
}
\caption{We show a few representative examples, description of the phenomenon the template is intended to test, top human annotated label, and associated confidence (0-1). T15 and T16 are examples of relatively \textit{harder} templates requiring multiple capabilities. T15 require comparative, negation, and syntactic. T16 requires commonsense and numerical reasoning.}
\label{tab:exampletemplates}
\end{table*}

\subsection{Template Generation for Reasoning Categories}

Automatic template creation from public datasets is not straightforward, as examples in public NLI datasets represent multiple capabilities \citep{joshi2020taxinli}. Even targeted datasets such as Winograd Schema Challenge \citep{levesque2012winograd} require careful re-annotation, as examples may require lexical or boolean reasoning.
Instead, we manually create templates and use human annotations to verify the correctness of templated instances. As and when required, we extend the list of basic key placeholders (and corresponding lexicon) provided by the \textsc{CheckList} tool, such as, \textcolor{darkblue}{\big\{PROFESSION\big\}} = \{doctor, actor, politician, $\ldots$\}, \textcolor{darkblue}{\big\{COM ADJ\big\}} = \{smarter, taller\, $\ldots$\}, \textcolor{darkblue}{\big\{CITY\big\}} = \{Paris, New York\, $\ldots$\}.\\
The process of template creation followed this general pattern. In step 1, we make a list of phenomenon that fall under a capability. For example, under \underline{Comparative}, the phenomenons are (a) checking if the predicates are comparable, (b) understanding of comparative adjectives (c) understanding of superlative adjectives and (d) mixed (both (b) and (c) in the same template). For \underline{Spatial}, the phenomenons are (a) 2D space reasoning (on, above, below, left, right), (b) direction (North, South, East, West) (c) comparison of distance. In step 2, there are multiple templates created to test the phenomenon. These templates are logical perturbations of each other. We vary the number of predicates, reorder the predicates, reorder the templated sentences in the premise (if there are more than one), or replace a number with its romanized form while retaining intended meaning. This leads to multiple templates generated for each of the phenomenon we listed in step 1. If a phenomenon is difficult to translate to a template, they are either ruled out or we use an external resource to aid the template creation. One author came up with the templates while the others verified the validity of the examples. The entire suite of templates required close to 40 hours for creation.
In the subsequent paragraphs, we highlight interesting templates (Tab.~\ref{tab:exampletemplates}) and discuss some challenges faced during template creation. 
\\
\textbf{Linguistic.}~~~The \underline{Lexical} templates test understanding of synonyms, antonyms, hypernyms, and unrelated phrases. 
\underline{Syntactic} templates test different forms of ellipsis (``P: Hugh is a lecturer and Nancy is too. H: Nancy is a lecturer."). Templates for Paraphrasing are hard to create because most paraphrases in existing-paraphrase corpora \citep{dolan2005automatically,WinNT} are not necessarily entailments, and paraphrases may involve lexical changes as well. 
\\\noindent
\textbf{Logical.}~~~ For \underline{Negation}, templates test the effect of negation words (not, never, hardly) in the premise or hypothesis. 
\underline{Boolean} templates test logical \textit{and} ($\land$), \textit{or} ($\lor$) and their combination; ordered resolution (T1, T2) of varying lengths along with inclusion of negation. 
\underline{Quantifier} templates test the understanding of \textit{universal} (all, $\forall$), and \textit{existential} (some, none; $\exists,\neg\exists$) quantification, and the effects of interchanging them (``P: None of the clips are brown in colour. H: Some of the clips are brown in colour.").
\underline{Comparative} templates range from basic comparative checks (can A and B be compared) to more complex comparisons, such as comparative, superlative, and associativity (``P: Peter is dumber than Diane. Diane is dumber than George. H: Peter is dumber than George."). We also include templates with insufficient information (``P: Among Albert, Larry and Susan the wealthiest is Larry. H: Susan is wealthier than Albert."). We collect a list of adjectives along with their comparative and superlative forms for these templates.
For \underline{Coreference}, we come up with templates to test gendered (T3) and animate vs. inanimate resolution (``P: Mike has a pencil. It is crimson in colour. H: Pencil is crimson in colour."). 
\underline{Spatial} templates test understanding of spatial prepositions and adverbs indicating relative positions such as near-far (``P: Irving is 350 miles from San Francisco and 570 miles from Miami. H: Irving is nearer to San Francisco than Irving is to Miami."), above-below, left-right (``P: Helen is to the left of Annie. Annie is to the left of Johnny. H: Johnny is to the right of Helen."). We also include a set of templates testing cardinal directions (north, east, south, and west); and some requiring comparison of distances (T4, requiring both Spatial and Numerical understanding).
\underline{Temporal} templates cover relative occurrences of events using prepositions, such as \textit{before-after} and \textit{earlier-later}. Another set of templates tests the understanding of time in the day (``P: Leslie reached hospital at 6 AM whereas Julie reached at 11:30 AM. H: Leslie reached before Julie.") , month or year. We add templates for temporally ordered events (T5) which require the inference of the earliest or latest events (similar to \cite{vashishtha-etal-2020-temporal}). 
\underline{Causal} template creation is tricky since often reasoning beyond \textit{form} is required. However, with specific controls placed, we can generate accurate causal premise-hypothesis pairs. We use a set of action-verb pairs which are complementary (e.g. give-take, lend-borrow, etc.) to describe corresponding actions between 2 entities and an object (T6). This still results in limited test cases. Explorations into leveraging knowledge graphs such as ConceptNet, ATOMIC \citep{speer-conceptnet,DBLP:conf/aaai/SapBABLRRSC19} to retrieve appropriate causal action phrases could be a way to tackle the issue.
Most \textsc{Logical} category templates implicitly test the \underline{Relational} (deductive reasoning with relations in text) capability, and hence we add three representative templates. \underline{Numerical} templates test a basic understanding of counting (T7, T8), addition, subtraction and numerical comparison (``P: Jeff has 2 coins. Don has 1 coins. H: Jeff has more coins than Don."). Following \cite{talmor2019olmpics}, we add another set of templates pertaining to counting.

\noindent
\textbf{Knowledge. }~~~For \underline{Taxonomic}, we create templates where a taxonomic hierarchy (external to text) is implicitly required to infer the hypothesis -- either requiring to infer A is a type of B (A has some properties. $\Rightarrow$ A is a type of B) or utilising that information (similar to \cite{NEURIPS2020_e992111e}) (B has property X $\Rightarrow$ A has property X). 
The scope of \underline{World} templates is vast. We add a representative template to test basic knowledge of geography (city-country pairs, T9)
\\\noindent
 \textbf{Pragmatic. }
For Pragmatic, we add templates along the lines of \cite{jeretic-etal-2020-natural}. \underline{Pre-supposition} templates test the existence of objects (T10), occurrence of events, aspectual verbs (T11) and quantifiers. Under \underline{Implicature} we create templates convering conventional implicature such as the use of word \textit{``even"} (``P: Even Martin bought a laptop. H: Martin was among the least likely to buy a laptop.") and generalized conversational implicature. For the latter, templates are created mainly following the Grice's cooperative principle of quantity \cite{grice1975logic}, also called scalar implicature (other Grice's cooperative principle include quality, relevance and manner). Most of the scalar implicature templates also require other capabilities such as quantifiers (T12), Boolean (T14), and Numerical (T15). Particularized conversational implicatures only hold true in limited context and thus it is difficult to create templates for them.

\subsection{\textsc{LoNLI} Dataset Statistics and Analysis} We created a total of 363 templates spanning all 17 capabilities. For each template, we generate 1000 examples by careful instantiations of the placeholders present in the template.  We ask two independent annotators, who are proficient in English and experienced in natural language annotation task to annotate the NLI label for 5 random examples from each template (1815 examples). The average Fleiss' $\kappa$ $0.81$ (same as Cohen's) shows very high inter-annotator agreement. The dataset can be found here \footnote{https://github.com/microsoft/lonli}.

\begin{figure}[!ht]
\centering
\subfloat{\includegraphics[width=0.6\textwidth]{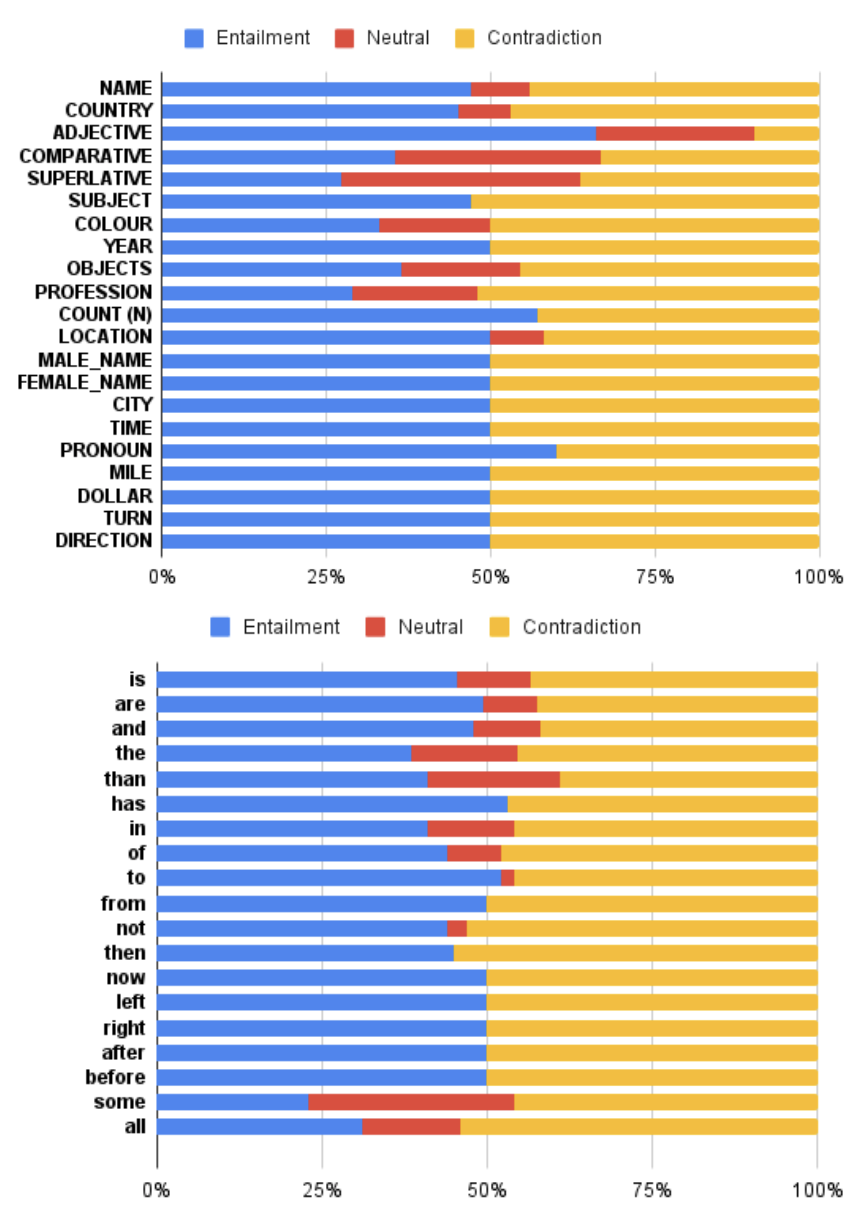}}
\caption{Lexical bias based on different keywords and top frequent words.}
\label{fig:histogramsbias}
\end{figure}

\paragraph{Analysing Artifacts and Bias in LoNLI}
We also analyze the lexical bias based on different keywords and top frequent words in the LoNLI dataset, show in Fig.~\ref{fig:histogramsbias}. In LoNLI, in most keyword dimensions, entailment and contradiction examples are balanced. As it is hard to synthetically and automatically create neutral examples, the percentage of neutral examples is limited in comparison to the other labels.

\section{LoNLI: Observations and Analysis}
\label{sec:observations}

The \textsc{LoNLI} dataset enables a multitude of analyses at different granularity. Through our analysis, we observe the following salient points: 1) MultiNLI fine-tuned systems fare poorly in most \textsc{Logical}, \textsc{Knowledge}, and \underline{implicature} categories. 2) In a post-hoc setting, due to these harder categories, the dataset remains  non-trivial to solve even for systems trained on additional targeted resources. 3) Most \textsc{Logical} categories require some unique information, that may not be learned solely from other capabilities. Apart from this, the framework of LoNLI allows us to attest to previously found observations and discover new phenomena-specific ones, such as systems oscillating inexplicably between logical and pragmatic interpretation for implicature examples. Intra-template variations help us find out model sensitivity towards gendered names and other lexicons (such as profession or adjectives).

\subsection{Analysis of NLI Systems}

We analyze four state-of-the-art systems -- BERT-base, DistilBERT-base, RoBERTa-large  \citep{devlin2019bert,liu2019roberta} and DeBERTa-large  \citep{he2020deberta} fine-tuned on MNLI data. Whenever available, we use the MNLI finetuned models from the Huggingface Transformers repository\footnote{\url{https://huggingface.co/models}}. Otherwise, we train it using the script from the repository \citep{Wolf2019HuggingFacesTS}, with the hyper-parameters suggested by corresponding papers.

\paragraph{Details of MNLI trained models.}
We use RoBERTa-large trained on MNLI from pytorch hub\footnote{\url{https://pytorch.org/hub/pytorch_fairseq_roberta/}} and DeBERTa-large trained on MNLI from huggingface model hub\footnote{\url{https://huggingface.co/microsoft/deberta-large-mnli}}. BERT and DistilBERT were not publicly available. So we train bert-base-uncased and distilbert-base-uncased with learning rate 5e-5, batch size 32, and sequence length 128 for 3 epochs\footnote{\url{https://github.com/huggingface/transformers/tree/master/examples/pytorch/text-classification}}. The validation was done after each epoch and the best model was saved.

\begin{figure*}[!ht]
	\centering
    \subfloat{\includegraphics[width=0.49\textwidth]{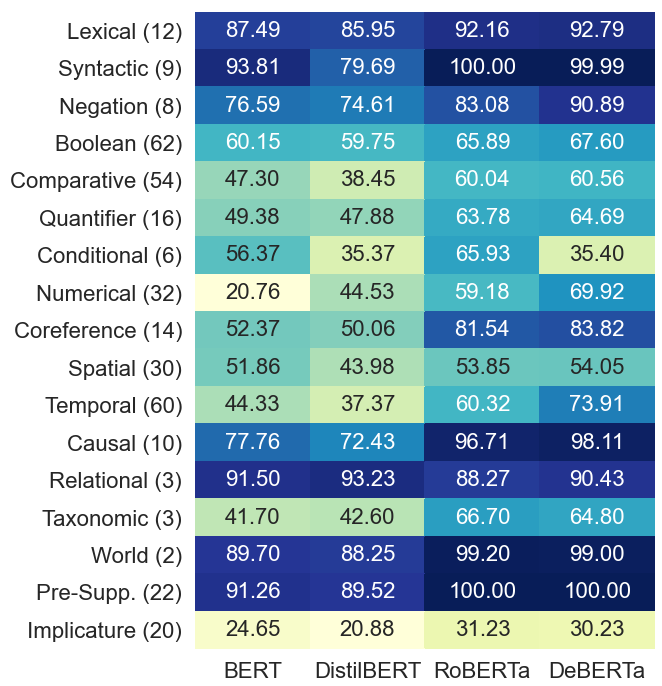}}
    \subfloat{\includegraphics[width=0.49\textwidth, height=0.24\textheight]{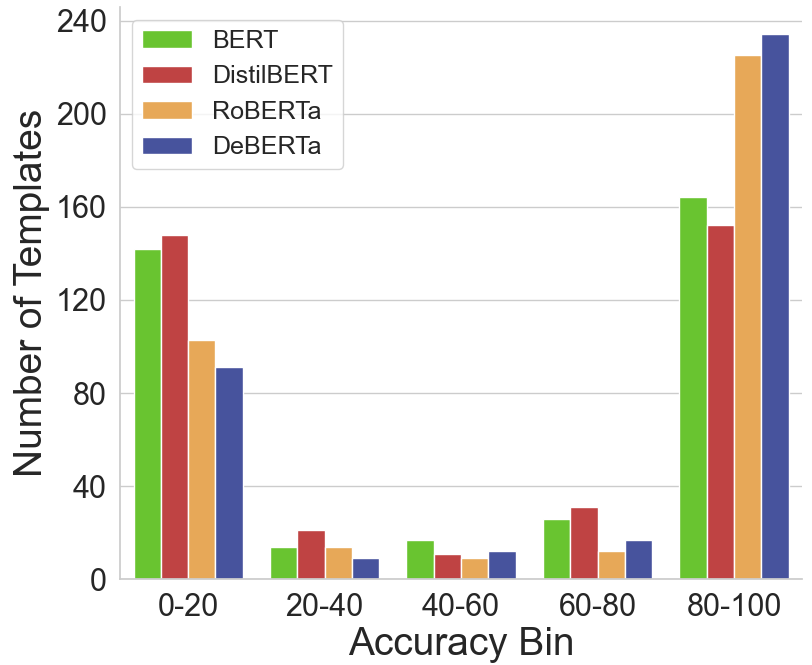}}
    \caption{(a) (Best viewed in color) For each of the 17 reasoning capabilities, we show average accuracy for each model (darker the color, higher the accuracy), (b) We show test-suite wide model accuracies divided into 5 bins. Here RoBERTa signifies, RoBERTa-large finetuned  on ANLI.
     }
\label{fig:app:histograms}
\end{figure*}

\begin{table}[!ht]
\centering
\resizebox{\columnwidth}{!}{%
\begin{tabular}{cccccc}
\toprule
Dataset & BERT & DistilBERT & RoBERTa & RoBERTa & DeBERTa \\
& & & MNLI & (M+S+F+A) & \\
\midrule
MNLI-test & 84.5 & 82.2 & 90.2 & 91.0 & \textbf{91.3/91.1} \\
\midrule
CheckList & 59.4 & 54.6 & 68.2 & \textbf{71.1} & 69.9 \\
\bottomrule
\end{tabular}%
}
\caption{Average Accuracies of all systems}
\label{tab:app:accuracy_all}
\end{table}
In Tab.~\ref{tab:app:accuracy_all}, we show accuracies of 5 state-of-the-art NLI systems. We observe the effect of adversarial training with more data using the RoBERTa-large trained on Adversarial NLI dataset \citep{nie2019adversarial} (primarily the round 3 model trained on MNLI, SNLI, Fever, and ANLI), and using a larger model DeBERTa-large \citep{he2020deberta} trained on Multi-NLI dataset which is at this point leader in the GLUE leaderboard. Interestingly, our capability-wise analysis in Figure.~\ref{fig:app:histograms}(a) shows that DeBERTa does only marginally better than RoBERTa-MNLI. It still suffers in spatial, numerical, knowledge, and implicature templates. In many cases, DeBERTa shows similar intra-template inconsistencies (Fig.~\ref{fig:app:histograms}(b)). RoBERTa-ANLI (M+S+F+A) model, only provides a $2.9\%$ accuracy improvement on \textsc{LoNLI}, showing the test suite is quite hard. Similar to DeBERTa, it suffers in spatial, Numerical, knowledge, implicature, and some more logical categories. However, we see from Fig.~\ref{fig:app:histograms}(b), that the intra-template inconsistencies decrease even more for RoBERTa-ANLI. Next, we analyze further at the level of i) capabilities, ii) templates, and iii) intra-template granularity.



\paragraph{Capability-wise Performance.}
From the capability-wise average accuracy (Figure \ref{fig:app:histograms}(a)), we observe that all models perform well on Lexical, Syntactic, and Presupposition capabilities. For the \textsc{Logical} categories, the results are comparably poor and inconsistent across both the capabilities and model dimensions. The same holds for the \textsc{Knowledge} categories and implicature templates. In short, a subset of \textsc{Logical} categories (\underline{comparative}, \underline{spatial}, \underline{temporal}, \underline{quantifier}, \underline{numerical}), \underline{implicature}, and \underline{taxonomic}  stands out as relatively \textit{harder} capabilities across systems. Through further template-wise analysis, we affirm many previous observations and discover new ones related to these harder capabilities.
For the ensuing analysis, we mark a template as \textit{passed} if the model's accuracy is above $80\%$, as \textit{unsure} if the accuracy is between $20$ to $80\%$, and \textit{failed} if less than $20\%$. 
 
 \paragraph{Reproducing Previous Observations.}
In \underline{temporal}, models are sensitive to lexical substitutions \citep{glockner-etal-2018-breaking} such as replacing ``before'' (``after'') with  ``earlier than'' (``later than''). The models show a label bias towards contradiction in the presence of ``not'' even if the true label is entailment \citep{naik-etal-2018-stress}.
Similar to \cite{Talmor2020oLMpicsOnWL}, in \underline{numerical}, we find that all models struggle in comparing age given year of birth. BERT and DistilBERT are unsure of most templates. RoBERTa and DeBERTa perform better at counting while failing when the hypothesis refers to an incorrect count.  All models struggle with templates pertaining to addition and subtraction (often showing label bias) \citep{Pikekos2021MeasuringAI, Wallace2019Numbers, naik-etal-2018-stress}. On \underline{quantifier} templates with ``some-all'' the result of BERT and DistilBERT are neither consistent with the pragmatic nor logical mode of reasoning \citep{jeretic-etal-2020-natural}. Failure on many templates (for example all models fail to predict contradiction on P: Sue but not Harry is a dentist. H: Harry is a dentist.) can be attributed to shallow, fallible syntactic heuristics rather than ``understanding'' \citep{mccoy-etal-2019-right}.

\paragraph{Uncovering New Findings.}
On the ordered resolution of varying lengths, we observe that 
RoBERTa and DeBERTa are biased towards entailment (\textit{unsure} for contradiction). Within \underline{comparative}, RoBERTa and DeBERTa fail on templates testing associativity of comparison (P: A is X than B, B is X than C. H: A is X than C). All models fail on templates where the information is insufficient (P: A is X than C, B is X than C H: A is X than B). BERT and DistilBERT are \textit{unsure} when the premise or hypothesis requires reasoning over a superlative adjective. 
On another set of templates for counting (P: Mary and Billy are the only children of Joe. H: Joe has exactly 2 children.), we find even RoBERTa and DeBERTa struggle which is in contrast to \cite{Talmor2020oLMpicsOnWL}. This indicates that RoBERTa and DeBERTa may not actually \textit{know} how to count, but rather may rely on spurious correlations.
Within \underline{temporal}, 
all models fail to detect the ``first'' or the ``last'' event in a sequence of three events (A then B then C). 
On \underline{implicature} templates, the models oscillate unexplainably between logical (P: Silverware and plate lie on the table. Barbara asked for the plate. H: Barbara also asked for the silverware.; predicting neutral over contradiction) and pragmatic interpretation (P: Some of the balls are purple in color. H: All of the balls are purple in color.; predicting contradiction over neutral), depending on the template.  \\

\subsection{Investigating Intra-Template Variations}

We also analyze \textbf{intra-template} variations, i.e., why some templates fall in the middle bin ($20$-$80\%$; Fig.~\ref{fig:app:histograms}(b)). The accuracy of a template lying in the middle bin can be explained using the choice of lexemes for the placeholders. Even here, we attest to previously observed model sensitivity towards gendered names \citep{NEURIPS2020_92650b2e, gender-bias}. We find that models are sensitive to dimensions of other lexicons as well such as profession or adjectives. We expand on our analysis of variation within a template arising due to lexicons by discussing two example templates.

\begin{table}[!ht]
\centering
\begin{tabular}{ccc}
\toprule
\{PROFESSION\} & Male & Female \\
\midrule
engineer       & 0.00    & 0.01      \\
poet           & 0.00    & 0.04      \\
entrepreneur   & 0.01    & 0.01      \\
politician     & 0.02    & 0.35     \\
writer         & 0.03    & 0.07      \\
banker         & 0.06    & 0.08      \\
dancer         & 0.9    & 0.12     \\
actor          & 0.10   & 0.26     \\
painter        & 0.10   & 0.33     \\
accountant     & 0.12   & 0.35     \\
businessman    & 0.16   & 0.32     \\
author         & 0.33   & 0.45     \\
singer         & 0.49   & 0.77     \\
teacher        & 0.62   & 0.63     \\
doctor         & 0.80   & 0.77     \\
professor      & 0.92   & 0.83    \\
\bottomrule
\end{tabular}%
\caption{Variation in accuracy in BERT for different values of professions and gender}
\label{tab:app:lexical_variation1}
\end{table}

\begin{table}[!ht]
\centering
\begin{tabular}{ccc}
\toprule
\{ADJ\} & Male & Female \\
\midrule
big     & 0.00    & 0.03      \\
sweet   & 0.16   & 0.11     \\
smart   & 0.25   & 0.20     \\
weird   & 0.38   & 0.40     \\
strong  & 0.41   & 0.40     \\
tough   & 0.47   & 0.48     \\
old     & 0.56   & 0.55     \\
tall    & 0.56   & 0.61     \\
tiny    & 0.99   & 0.99     \\
creepy  & 1.00  & 1.00   \\
\bottomrule
\end{tabular}
\caption{Variation in the accuracy of BERT for different values of adjective and gender}
\label{tab:app:lexical_variation2}
\end{table}

\paragraph{Case Study 1 on PROFESSION.} 
The expected label for the template {\tt P: \{NAME1\}, but not \{NAME2\}, is a \{PROFESSION\}.} {\tt H: \{NAME2\} is a \{PROFESSION\}. } is ``contradiction''.
We create examples from this template by varying on two dimensions, 1) first we vary the gender of \textcolor{darkblue}{NAME2} (we keep \textcolor{darkblue}{NAME1} to be the same gender as \textcolor{darkblue}{NAME2} to avoid any confusion), and then 2) use different values of the \textcolor{darkblue}{PROFESSION} lexicon, and then vary only names. For each combination of the above two we sample 100 examples and report BERT's accuracy in Table \ref{tab:app:lexical_variation1}. As mentioned, accuracy varies strongly with the change in the lexicon for \textcolor{darkblue}{PROFESSION}. Template accuracy varies from low when the profession is set to engineer ($0\%$), poet ($0\%$) to high when the profession is set to doctor ($82\%$), professor ($92\%$).

\paragraph{Case Study 2 on Adjectives.} 
Next, we select another template: {\tt P: \{NAME1\} is \{ADJ\}. \{NAME2\} is \{COM ADJ\}.}{\tt H: \{NAME2\} is \{COM ADJ\} than \{NAME1\}.} For this template, expected label is entailment.
Similar to before, we generate examples conditioned on \textcolor{darkblue}{ADJ} and gender. Table \ref{tab:app:lexical_variation2} shows the accuracy for different combinations of gender and \textcolor{darkblue}{ADJ}. Compared to the previous template, we notice that the accuracies are almost unchanged across the gender dimension. This means that compared to \textcolor{darkblue}{ADJ}, BERT shows bias towards certain professions \textcolor{darkblue}{PROFESSION} conditioned on gender. Here also template accuracy varies from low when adjectives are bigger ($0\%$), sweeter ($16\%$) to  high when adjectives are tiny ($99\%$), creepier ($100\%$).

\subsection{Effect of Existing Resources}

From Tab.~\ref{tab:app:accuracy_all}, we observed that current SOTA models perform poorly on \textsc{LoNLI}, and found certain capabilities compratively \textit{harder}. Venturing further to assess the \textit{hardness}, we examine whether further finetuning on specific existing datasets can help. We select two capability-aware resources, CLCD (aimed at \textit{connectives}), and \textsc{ImpPres} (aimed at \textsc{Pragmatics}), and an adversarial resource ANLI. We use MNLI-trained BERT and RoBERTa, and further train them on the three resources. 

\paragraph{Finetuning on additional resources.}
CLCD dataset is a two-way classification task -- contradiction and non-contradiction. So, first, we convert the examples to traditional 3-way labeling tasks by labeling non-contradiction with neutral and entailment. As CLCD is synthetic (and templated) in nature, we use the templates to re-label sets of examples than individual ones. Also CLCD and \textsc{ImpPres} do not provide a separate test split so we sample 10\% data for testing and remove it from training. ANLI has three subsets R1, R2, and R3 which we use in a combined fashion by concatenating the datasets. For fine-tuning, we start with BERT-base and RoBERTa-large fine-tuned on MNLI and train them on these three datasets using a learning rate of 5e-5. Since the datasets are smaller, we experiment with smaller batch sizes of 4 and 8. For CLCD and \textsc{ImpPres} the sequence length remains 128 but for ANLI we increase the sequence length to 256 (following \cite{nie2019adversarial}). For both, we train the model for 3 epochs and save the model with the best validation score after each epoch.

\begin{figure}[!ht]
	\centering
    \subfloat{\includegraphics[width=0.7\textwidth]{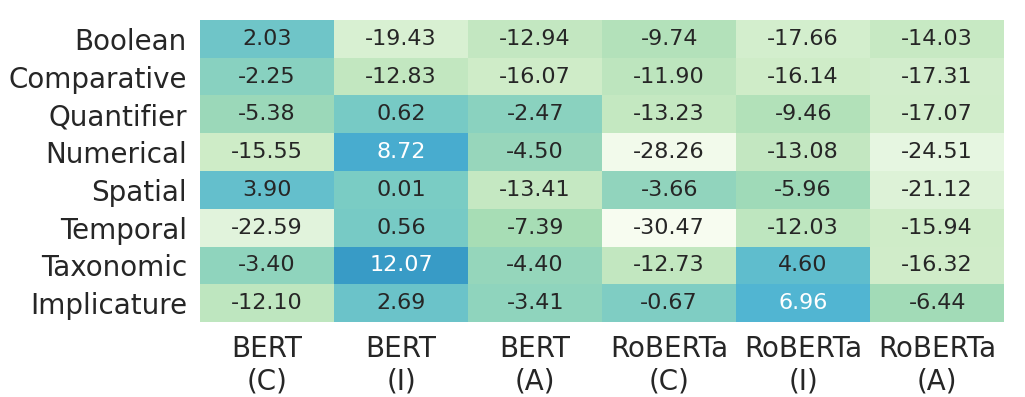}}\hfill
    \caption{We show the increment in accuracy compared to MNLI variants, for select capabilities. (C)=CLCD, (I)=\textsc{ImpPres}, (A)=ANLI.}
\label{fig:heatmap_existing_resources}
\end{figure}

\paragraph{Results.}
Fig.~\ref{fig:heatmap_existing_resources} summarizes the capability-wise accuracy of the six variants on selected capabilities of interest. Expectedly, training on \textsc{ImpPres} shows benefit (albeit marginal) on \underline{implicature} for both models over their MNLI-variants. \textsc{ImpPres} benefits BERT on most capabilities. For RoBERTa, most other capabilities suffer except for \underline{taxonomic}. Training on CLCD benefits BERT marginally on \underline{boolean} capability, while preserving its performance on \underline{comparative} and \underline{quantifier}. Even here, improvements for RoBERTa are minimal. We also see that continued finetuning on capability-aware datasets is often detrimental to the rest of the capabilities. \underline{Numerical} capability suffers the most on both datasets, while \underline{temporal} suffers the most on CLCD. ANLI is capability-agnostic and contains harder examples. We hardly observe any benefit of using both MNLI and ANLI for both models (all capabilities suffer marginally). For ANLI, the decline in performance is higher for BERT than RoBERTa. In summary, we observe that \textsc{LoNLI} is non-trivial to solve even with additional resources.

   
\subsection{Few-shot and Cross-capability Finetuning} 
  Finally, we use the \textsc{LoNLI} dataset (as train-test) to investigate whether the \textit{harder} capabilities can be learned from others (in a leave-one-out setting) or in a few-shot manner. 
  
   \begin{figure}[!ht]
    \centering
    \includegraphics[width=0.9\columnwidth]{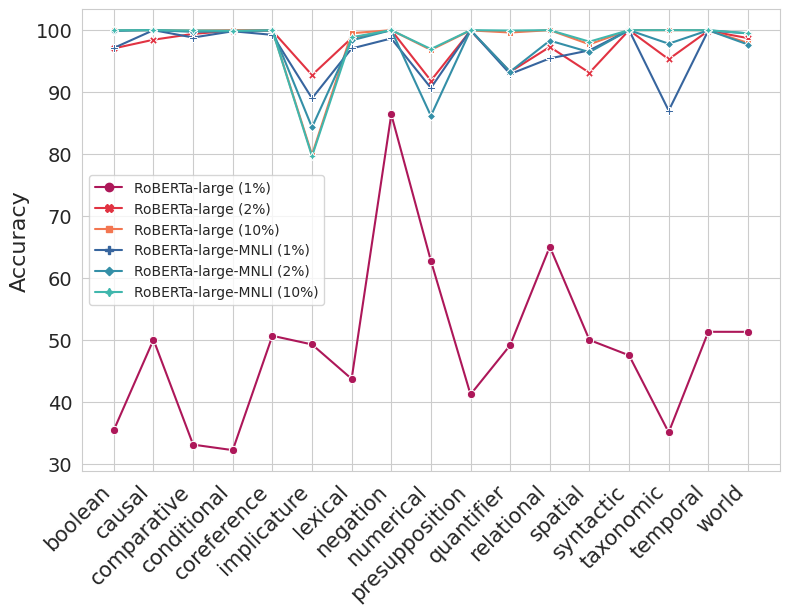}
    \caption{(Few-shot) Figure shows how RoBERTa-large and RoBERTa-large-MNLI performance varies with $1-10\%$ of \textsc{LoNLI} training data. For more than $10\%$, the accuracy varies minimally.} 
    \label{fig:fewshot}
\end{figure}
  \paragraph{Few-shot Finetuning.} We employ few-shot finetuning to gauge if some categories remain harder with the addition of data. We uniformly sample training data for each capability, then gradually increase training data from $1\%, 2\%, 10\%, 20\%$ to $30\%$ for each experiment. Using the sampled data, we finetune pre-trained RoBERTa-large and RoBERTa-large trained on MultiNLI\footnote{For each model, we perform a hyperparameter search using different learning-rate ($1e-5, 2e-5, 5e-5$), warm-up steps ($0, 500$), scheduler (linear, constant), and training epoch ($3, 5$) combinations; and choose  the hyperparameters that provide the highest LoNLI and MultiNLI validation accuracy.}.  We test on $20\%$ of the dataset. As shown in Figure \ref{fig:fewshot},  pre-trained RoBERTa-large performs very low on all categories with $1\%$ training data but quickly reaches near 100\% accuracy with $2\%$ data. As expected, even with $1\%$ data, RoBERTa-large-MNLI achieves 100\% accuracy on all categories except \underline{implicature}, \underline{numerical}, \underline{quantifier} and \underline{taxonomic}.
  \\\noindent
  \textbf{The curious case of implicature.} Some implicature templates stand out as the outcome may vary based on whether pragmatic or logical interpretation is employed. Interestingly, RoBERTa-large-MNLI becomes somewhat consistent in adapting logical interpretations for templates that may have different outcomes for different interpretations. This is expected, as without additional input context about which interpretation is expected, the model should be consistent.
  
  \paragraph{Cross-capability Finetuning.} Next we design leave-one-out (\texttt{loo}) and bring-one-in (\texttt{boi}) experiments. For \texttt{loo}, we remove a capability altogether from training and train using $80\%$ of the remaining data (train/val/test: 64/16/20). For comparison, we create the \texttt{loo-all} setting where no capability is left out. For the dual setting \texttt{boi}, we train using $80\%$ of one particular capability (leaving others out). For both, we test using $20\%$ of the entire dataset.
  \begin{figure}[!ht]
    \centering
    \includegraphics[width=0.9\columnwidth]{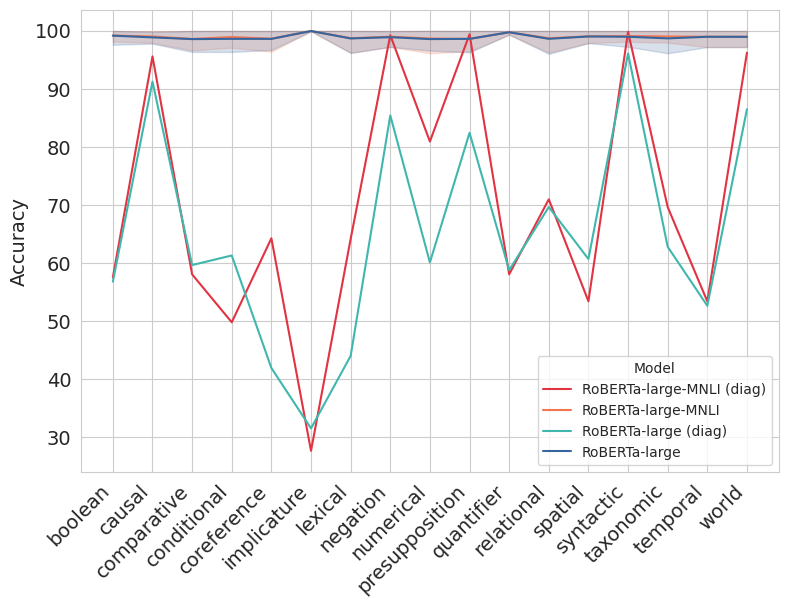}
    \caption{(Cross-capability \texttt{loo}) Capability-wise accuracies after RoBERTa-large, and RoBERTa-large-MNLI finetuning when an individual (X-axis) capability is left out. The \textit{diag} indicates the test accuracies of the left-out capabilities, for example, both models achieve 57\% accuracy for boolean (and close to 99\% for others) when boolean is left out.}
    \label{fig:crosscap}
\end{figure}
\begin{figure}[!ht]
    \centering
    \includegraphics[width=0.9\columnwidth]{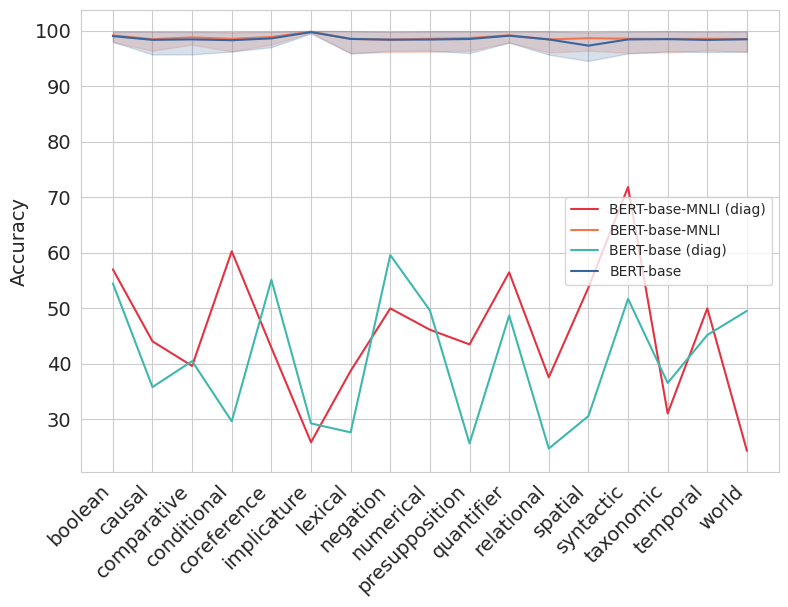}
    \caption{(Cross-capability \texttt{loo}) Capability-wise accuracies after BERT, and BERT finetuning when an individual (X-axis) capability is left out. The \textit{diag} indicates the test accuracies of the left-out capabilities, for example, BERT-base-MNLI models achieve 56\% accuracy for boolean (and close to 99\% for others) when boolean is left out. }
    \label{fig:crosscap_bert}
\end{figure}
   We fine-tune and plot RoBERTa-large and RoBERTa-large-MNLI accuracies in the \texttt{loo} setting in Figure \ref{fig:crosscap} (BERT results in Figure~\ref{fig:crosscap_bert}). The X-axis denotes the left-out capabilities (i.e., withheld during training). The Y-axis denotes test accuracy per capability (left-out capability's test accuracy for \textit{diag}, average with error bars for others). We observe the following patterns. 
  \\\noindent
  \textbf{Can capabilities be acquired by training others?}  For both models (in \texttt{loo}), the average test performance of the left-out capabilities remains low (RB-MNLI: $70.5\pm 20.7$, RB: $64.8\pm16.64$), showing most capabilities (and our selected templates) need some unique reasoning or information not provided by others. The capabilities comparative, conditional, coreference, implicature, quantifier, spatial, and temporal show the most degradation (compared to \texttt{loo-all}, where examples from all capabilities are used during training).
  \\\noindent 
  \textbf{Are there any spurious correlations?} Leaving most capabilities seem to improve quantifier and implicature performance (compared to \texttt{loo-all} scenario). While implicature performance can be explained, we experiment with RoBERTa-large-MNLI in \texttt{boi} setting where we train on one capability and test on others. Other than lexical, numerical, implicature and spatial, each category individually improves on quantifier ($3.4 \pm 10.7$) compared to RoBERTa-large-MNLI model. A very similar, but prominent improvement happens for numerical ($12.94 \pm 11.78$), and spatial ($10.36 \pm 14.24$). 
  Interestingly, for spatial and numerical, there is hardly any relevant information in other templates; showing how easily these models pick up unintended spurious patterns.
  \\\noindent
  \textbf{Does MNLI Finetuning \textit{help} acquire the harder capabilities?} For RoBERTa-large and BERT-base (in Fig.~\ref{fig:crosscap}, \ref{fig:crosscap_bert}), finetuning on MNLI followed by other categories do not seem to help pick up comparative, quantifier, and implicature. From both models, we observe that for comparative, quantifier, and implicature; the left-out capability's accuracy (in \texttt{loo}, see \textit{diag} in Fig.~\ref{fig:crosscap}) often degrades when we use MNLI-finetuned model, compared to the pre-trained model. This underscores the necessity for category-specific evaluation and enhancement of NLI systems.

\section{Conclusion}
\vspace{-0.05in}
Using the \textsc{CheckList} framework, we propose an extensible semi-synthetic NLI benchmark ($363$ templates, $363$k examples) with reasoning type annotations, covering 17 reasoning types (extending a recently proposed NLI taxonomy).  Unlike other NLI datasets, our dataset 1) comes with explicit reasoning type annotations covering a wide range of linguistic and logical phenomena, 2) enables fine-grained capability-wise analysis, 3) is easily extensible in the future, and 4) is devoid of known artifacts and bias.  SOTA NLI systems perform poorly in this dataset, even with capability-specific existing resources as additional training data. Further analysis leads us to find a subset of capabilities that are harder across systems. Through fine-tuning experiments, we find some remain harder even with the addition of training data from the test-suite, and are hard to learn solely from other capabilities.

\appendix

\section{Benchmarking: Detailed Observations from Template Perturbations}
We provide a list of interesting templates in Table \ref{tab:app:interestingtemplates}, from where we can glean more fine-grained observations about the underlying systems' behavior. \\
{\tt \underline{T2}:} All models fails on simple boolean template for testing ``or''.\\
{\tt \underline{T13,T14}:} BERT and DistilBERT are {\it unsure} on ordered resolution. RoBERTa is able to predict accurately when the label is entailment but {\it unsure} when the label is a contradiction. The observation remains consistent for chains of length 2, 3, and 4. \\
{\tt \underline{T21,T22}:} We modify {\tt T13,T14} by introducing ``not'' in the hypothesis and observe a label shift towards contradiction for all models (even RoBERTa which was accurate for entailment templates).\\
{\tt \underline{T45,T46}:} We look at both these templates together and observe that BERT and DistilBERT are biased towards entailment labels while not understanding gendered names. RoBERTa on the other hand performs fairly accurately on both templates. \\
{\tt \underline{T63}:} DistilBERT fails on this very basic comparative template where the arguments are swapped.\\
{\tt \underline{T68,T71}:} The information within the premise is not sufficient to arrive at a hypothesis. All models struggle with templates such as this.\\
{\tt \underline{T76,T77}:} This template requires comparative and syntactic understanding. RoBERTa performs accurately on both these templates whereas BERT and DistilBERT are {\it unsure}. On further analysis, we observe that BERT has a lexical bias for the placeholder \textcolor{darkblue}{ADJ}. \\
{\tt \underline{T80,T81}:} All models fail on template related to 2D directions.\\
{\tt \underline{T88,T89,T92,T93}:} All models are {\it unsure} or biased on this set of templates. Since RoBERTa is able to perform numerical comparison we would expect it to compare years but that is not observed. An interesting observation is BERT being sensitive to the lexical substitution ``before'' (or ``after'') to ``earlier than`` (or ``later than'').\\
{\tt \underline{T98,T99}:} RoBERTa is able to correctly reason out the relative ordering to events whereas BERT and DistilBERT are {\it unsure}\\
{\tt \underline{T116,T117}:} Compared to BERT and DistilBERT, RoBERTa is better at understanding causal-verb pairs.\\
{\tt \underline{T122}:} A very simple quantifier template on which both BERT and DistilBERT fail.\\
{\tt \underline{T171,T172}:} These two are exactly the same template with different label depending on ``logical'' vs ``implicative'' reasoning. BERT and DistilBERT are {\it unsure} whereas RoBERTa is ``logical'' .\\
{\tt \underline{T128,T127}:} Another pair of exactly same template but this time all the three models predict contradiction which is ``implicative'' reasoning.

\begin{table*}[!ht]
\scriptsize
\setlength\fboxsep{1pt}

\resizebox{\textwidth}{!}{
\begin{tabular}{p{0.6cm} p{12.5cm} ccc}
\toprule
\# & \multicolumn{1}{c}{\multirow{2}{*}{Template}} & \multicolumn{3}{c}{Performance (Accuracy \%)}\\ \cline{3-5}\noalign{\smallskip}
& & \multicolumn{1}{c}{\tiny BERT} & \multicolumn{1}{c}{\tiny DistilBERT} & \multicolumn{1}{c}{\tiny RoBERTa} \\ 

\arrayrulecolor{black}\midrule
T2 & {\textcolor{premise}P:} \{NAME1\} or \{NAME2\} is \{ADJ\}. {\textcolor{hypothesis}H:} \{NAME1/2\} is \{ADJ\}. {\colorbox{pink}{\texttt{neutral}}} & \multirow{2}{*}{1.80} & \multirow{2}{*}{0.00} & \multirow{2}{*}{1.60}\\
Bool & {\textcolor{premise}P:} Ann or Barbara is optimistic. {\textcolor{hypothesis}H:} Ann is optimistic\\
\midrule
T13 & {\textcolor{premise}P:} \{NAME1\} and \{NAME2\} are from \{CTRY1\} and \{CTRY2\} respectively. {\textcolor{hypothesis}H:} \{NAME1\} is from \{CTRY1\}. {\colorbox{pink}{\texttt{entail}}} & \multirow{2}{*}{57.10} & \multirow{2}{*}{59.20} & \multirow{2}{*}{100.00}\\
Bool & {\textcolor{premise}P:} Mary and David are from Canada and Australia respectively. {\textcolor{hypothesis}H:} Mary is from Canada.\\
\midrule
T14 & {\textcolor{premise}P:} \{NAME1\} and \{NAME2\} are from \{CTRY1\} and \{CTRY2\} respectively. {\textcolor{hypothesis}H:} \{NAME1\} is from \{CTRY2\}. {\colorbox{pink}{\texttt{cont}}} & \multirow{2}{*}{46.00} & \multirow{2}{*}{39.30} & \multirow{2}{*}{49.50}\\
Bool & {\textcolor{premise}P:} Robert and Charles are from Russia and France respectively. {\textcolor{hypothesis}H:} Charles is from Russia.\\
\midrule
T21 & {\textcolor{premise}P:} \{NAME1\} and \{NAME2\} are from \{CTRY1\} and \{CTRY2\} respectively. {\textcolor{hypothesis}H:} \{NAME1\} is not from \{CTRY2\}. {\colorbox{pink}{\texttt{entail}}} & \multirow{2}{*}{0.00} & \multirow{2}{*}{0.00} & \multirow{2}{*}{0.00}\\
Bool & {\textcolor{premise}P:} Margaret and Robert are from America and Russia respectively. {\textcolor{hypothesis}H:} Margaret is not from Russia.\\
\midrule
T22 & {\textcolor{premise}P:} \{NAME1\} and \{NAME2\} are from \{CTRY1\} and \{CTRY2\} respectively. {\textcolor{hypothesis}H:} \{NAME1\} is not from \{CTRY1\}. {\colorbox{pink}{\texttt{cont}}} & \multirow{2}{*}{100.00} & \multirow{2}{*}{100.00} & \multirow{2}{*}{100.00}\\
Bool & {\textcolor{premise}P:} John and James are from India and China respectively. {\textcolor{hypothesis}H:} James is not from China.\\
\midrule
T45 & {\textcolor{premise}P:} \{MALE NAME\} and \{FEMALE NAME\} are \{friends/collegues/married\}. He is \{a/an\} \{PROF1\} and she is \{a/an\} \{PROF2\}. {\textcolor{hypothesis}H:} \{MALE NAME\} is \{a/an\} \{PROF1\}. {\colorbox{pink}{\texttt{entail}}} & \multirow{2}{*}{99.50} & \multirow{2}{*}{98.00} & \multirow{2}{*}{97.70}\\
Coref & {\textcolor{premise}P:} Marlen and Rudolf work together. He is a minister and she is a professor. {\textcolor{hypothesis}H:} Rudolf is minister.\\
\midrule
T46 & {\textcolor{premise}P:} \{MALE NAME\} and \{FEMALE NAME\} are \{friends/collegues/married\}. He is \{a/an\} \{PROF1\} and she is \{a/an\} \{PROF2\}. {\textcolor{hypothesis}H:} \{MALE NAME\} is \{a/an\} \{PROF2\}. {\colorbox{pink}{\texttt{cont}}} & \multirow{2}{*}{4.70} & \multirow{2}{*}{2.40} & \multirow{2}{*}{92.60}\\
Coref & {\textcolor{premise}P:} Mujtaba and Teresa are colleagues. She is a farmer and he is a professor. {\textcolor{hypothesis}H:} Teresa is professor.\\
\midrule
T63 & {\textcolor{premise}P:} \{NAME1\} is \{COM ADJ\} than \{NAME2\} {\textcolor{hypothesis}H:} \{NAME2\} is \{COM ADJ\} than \{NAME1\}. {\colorbox{pink}{\texttt{cont}}} & \multirow{2}{*}{85.30} & \multirow{2}{*}{0.00} & \multirow{2}{*}{100.00}\\
Comp & {\textcolor{premise}P:} Elizabeth is harsher than Caroline. {\textcolor{hypothesis}H:} Caroline is harsher than Elizabeth.\\
\midrule
T68 & {\textcolor{premise}P:} \{NAME1\} is \{COM ADJ\} than \{NAME2\}. \{NAME1\} is \{COM ADJ\} than \{NAME3\}. {\textcolor{hypothesis}H:} \{NAME2/3\} is \{COM ADJ\} than \{NAME3/2\}. {\colorbox{pink}{\texttt{neutral}}} & \multirow{2}{*}{0.00} & \multirow{2}{*}{0.00} & \multirow{2}{*}{0.00}\\
Comp & {\textcolor{premise}P:} Philip is more important than Frances. Philip is more important than Kevin. {\textcolor{hypothesis}H:} Kevin is more important than Frances.\\
\midrule
T71 & {\textcolor{premise}P:} Among \{NAME1\}, \{NAME2\} and \{NAME3\} the \{SUP ADJ\} is \{NAME1\} {\textcolor{hypothesis}H:} \{NAME3/2\} is \{COM ADJ\} than \{NAME2/3\}. {\colorbox{pink}{\texttt{neutral}}} & \multirow{2}{*}{16.80} & \multirow{2}{*}{1.70} & \multirow{2}{*}{0.00}\\
Comp & {\textcolor{premise}P:} Among Dorothy, Peter, and Laura the healthiest is Peter. {\textcolor{hypothesis}H:} Dorothy is healthier than Laura.\\
\midrule
T76 & {\textcolor{premise}P:} \{NAME1\} is \{ADJ\}. \{NAME2\} is \{COM ADJ\}. {\textcolor{hypothesis}H:} \{NAME2\} is \{COM ADJ\} than \{NAME1\} {\colorbox{pink}{\texttt{entail}}} & \multirow{2}{*}{56.30} & \multirow{2}{*}{93.70} & \multirow{2}{*}{100.00}\\
Comp & {\textcolor{premise}P:} Emily is poor. Nancy is poorer. {\textcolor{hypothesis}H:} Nancy is poorer than Emily.\\
\midrule
T77 & {\textcolor{premise}P:} \{NAME1\} is \{ADJ\}. \{NAME2\} is \{COM ADJ\}. {\textcolor{hypothesis}H:} \{NAME1\} is \{COM ADJ\} than \{NAME2\} {\colorbox{pink}{\texttt{cont}}} & \multirow{2}{*}{45.90} & \multirow{2}{*}{0.00} & \multirow{2}{*}{88.90}\\
Comp & {\textcolor{premise}P:} Anna is smart. Julia is smarter. {\textcolor{hypothesis}H:} Anna is smarter than Julia.\\
\midrule
T80 & {\textcolor{premise}P:} \{NAME\} was facing \{DIR\} and turned towards his/her \{left/right/back\} {\textcolor{hypothesis}H:} \{NAME\} is now facing \{DIR\} {\colorbox{pink}{\texttt{entail}}} & \multirow{2}{*}{0.00} & \multirow{2}{*}{0.00} & \multirow{2}{*}{0.00}\\
Spatial & {\textcolor{premise}P:} Michael was facing east and turned towards his back. {\textcolor{hypothesis}H:} Michael is now facing west.\\
\midrule
T81 & {\textcolor{premise}P:} \{NAME\} was facing \{DIR\} and turned towards his/her \{left/right/back\} {\textcolor{hypothesis}H:} \{NAME\} is now facing \{DIR\} {\colorbox{pink}{\texttt{cont}}} & \multirow{2}{*}{30.20} & \multirow{2}{*}{30.20} & \multirow{2}{*}{0.40}\\
Spatial & {\textcolor{premise}P:} Jane was facing north and turned towards her back. {\textcolor{hypothesis}H:} Jane is now facing north.\\
\midrule
T82 & {\textcolor{premise}P:} \{CITY1\} is \{N1\} miles from \{CITY2\} and \{N2\} miles from \{CITY3\} {\textcolor{hypothesis}H:} \{CITY1\} is \{nearer/farther\} to \{CITY2\} than \{CITY3\} {\colorbox{pink}{\texttt{entail}}} & \multirow{2}{*}{99.50} & \multirow{2}{*}{82.20} & \multirow{2}{*}{88.90}\\
Spatial & {\textcolor{premise}P:} Hartford is 99 miles from Phoenix and 45 miles from Philadelphia. {\textcolor{hypothesis}H:} Hartford is nearer to Philadelphia than Phoenix.\\
\midrule
T83 & {\textcolor{premise}P:} \{CITY1\} is \{N1\} miles from \{CITY2\} and \{N2\} miles from \{CITY3\} {\textcolor{hypothesis}H:} \{CITY1\} is \{nearer/farther\} to \{CITY2\} than \{CITY3\} {\colorbox{pink}{\texttt{cont}}} & \multirow{2}{*}{0.80} & \multirow{2}{*}{6.70} & \multirow{2}{*}{49.70}\\
Spatial & {\textcolor{premise}P:} San Diego is 82 miles from Sacramento and 27 miles from Boston. {\textcolor{hypothesis}H:} San Diego is nearer to Sacramento than Boston.\\
\midrule
T88 & {\textcolor{premise}P:} \{NAME1\} was born in \{YEAR1\} and \{NAME2\} was born in \{YEAR2\}. {\textcolor{hypothesis}H:} \{NAME1/2\} was born \{before/after\} \{NAME1/2\} {\colorbox{pink}{\texttt{entail}}} & \multirow{2}{*}{2.70} & \multirow{2}{*}{41.50} & \multirow{2}{*}{95.60}\\
Temp & {\textcolor{premise}P:} Grace was born in 2004 and Susan was born in 1998. {\textcolor{hypothesis}H:} Susan was born before Grace.\\
\midrule
T89 & {\textcolor{premise}P:} \{NAME1\} was born in \{YEAR1\} and \{NAME2\} was born in \{YEAR2\}. {\textcolor{hypothesis}H:} \{NAME1/2\} was born \{before/after\} \{NAME1/2\} {\colorbox{pink}{\texttt{cont}}} & \multirow{2}{*}{95.80} & \multirow{2}{*}{35.60} & \multirow{2}{*}{0.00}\\
Temp & {\textcolor{premise}P:} Emma was born in 2016 and Harry was born in 1983. {\textcolor{hypothesis}H:} Harry was born after Emma.\\
\midrule
T92 & {\textcolor{premise}P:} \{NAME1\} was born in \{YEAR1\} and \{NAME2\} was born in \{YEAR2\}. {\textcolor{hypothesis}H:} \{NAME1/2\} was born \{earlier/later\} than \{NAME1/2\} {\colorbox{pink}{\texttt{entail}}} & \multirow{2}{*}{93.00} & \multirow{2}{*}{98.80} & \multirow{2}{*}{99.90}\\
Temp & {\textcolor{premise}P:} Christine was born in 2016 and Marie was born in 1985. {\textcolor{hypothesis}H:} Marie was born earlier than Christine.\\
\arrayrulecolor{black}\bottomrule
\end{tabular} 
}
\caption{We show some interesting Templates, and model accuracies on the templates.}
\label{tab:app:interestingtemplates}
\end{table*}

\begin{table*}[!ht]
\scriptsize
\setlength\fboxsep{1pt}
\resizebox{\textwidth}{!}{
\begin{tabular}{p{0.6cm} p{12.5cm} ccc}
\toprule
\# & \multicolumn{1}{c}{\multirow{2}{*}{Template}} & \multicolumn{3}{c}{Performance (Accuracy \%)}
\\
\cline{3-5}\noalign{\smallskip}
& & \multicolumn{1}{c}{\tiny BERT} & \multicolumn{1}{c}{\tiny DistilBERT} & \multicolumn{1}{c}{\tiny RoBERTa} \\ 
\midrule
T93 & {\textcolor{premise}P:} \{NAME1\} was born in \{YEAR1\} and \{NAME2\} was born in \{YEAR2\}. {\textcolor{hypothesis}H:} \{NAME1/2\} was born \{earlier/later\} than \{NAME1/2\} {\colorbox{pink}{\texttt{cont}}} & \multirow{2}{*}{0.20} & \multirow{2}{*}{0.10} & \multirow{2}{*}{0.00}\\
Temp & {\textcolor{premise}P:} Mark was born in 2007 and Mike was born in 1987. {\textcolor{hypothesis}H:} Mark was born earlier than Mike.\\
\midrule
T98 & {\textcolor{premise}P:} \{NAME\} has \{EVENT1\} followed by \{EVENT2\}. {\textcolor{hypothesis}H:} \{EVENT1/2\} is \{before/after\} \{EVENT1/2\} {\colorbox{pink}{\texttt{entail}}} & \multirow{2}{*}{50.00} & \multirow{2}{*}{35.40} & \multirow{2}{*}{88.60}\\
Temp & {\textcolor{premise}P:} Karen has lunch followed by a history class. {\textcolor{hypothesis}H:} The history class is after lunch.\\
\midrule
T99 & {\textcolor{premise}P:} \{NAME\} has \{EVENT1\} followed by \{EVENT2\}. {\textcolor{hypothesis}H:} \{EVENT1/2\} is \{before/after\} \{EVENT1/2\} {\colorbox{pink}{\texttt{cont}}} & \multirow{2}{*}{64.20} & \multirow{2}{*}{86.30} & \multirow{2}{*}{96.00}\\
Temp & {\textcolor{premise}P:} Kate has lunch followed by a meeting. {\textcolor{hypothesis}H:} The meeting is before lunch.\\
\midrule
T116 & {\textcolor{premise}P:} \{NAME1\} \{bought/taught/...\} \{OBJ\} to \{NAME2\} {\textcolor{hypothesis}H:} \{NAME2\} \{sold/learnt/...\} \{OBJ\} from \{NAME1\} {\colorbox{pink}{\texttt{entail}}} & \multirow{2}{*}{85.20} & \multirow{2}{*}{100.00} & \multirow{2}{*}{89.40}\\
Causal & {\textcolor{premise}P:} Jennifer taught physics to Rachel. {\textcolor{hypothesis}H:} Rachel learnt physics from Jennifer.\\
\midrule
T117 & {\textcolor{premise}P:} \{NAME1\} \{bought/taught/...\} \{OBJ\} to \{NAME2\} {\textcolor{hypothesis}H:} \{NAME1\} \{sold/learnt/...\} \{OBJ\} from \{NAME2\} {\colorbox{pink}{\texttt{cont}}} & \multirow{2}{*}{24.00} & \multirow{2}{*}{1.70} & \multirow{2}{*}{71.50}\\
Causal & {\textcolor{premise}P:} Barbara lent money to Laura. {\textcolor{hypothesis}H:} Barbara borrowed money from Laura.\\
\midrule
T122 & {\textcolor{premise}P:} None the \{OBJS\} are \{COLOR\} in color. {\textcolor{hypothesis}H:} Some of the \{OBJS\} are \{COLOR\} in color. {\colorbox{pink}{\texttt{cont}}} & \multirow{2}{*}{0.00} & \multirow{2}{*}{0.00} & \multirow{2}{*}{100.00}\\
Quan & {\textcolor{premise}P:} None of the wallpapers are green in colour. {\textcolor{hypothesis}H:} Some of the wallpapers are green in colour.\\
\midrule
T171 & {\textcolor{premise}P:} \{OBJ1\} and \{OBJ2\} lie on the table. \{NAME\} asked for \{OBJ1\} {\textcolor{hypothesis}H:} \{NAME\} also asked for \{OBJ2\} {\colorbox{pink}{\texttt{cont}}} & \multirow{2}{*}{0.00} & \multirow{2}{*}{0.00} & \multirow{2}{*}{0.00}\\
Impl & {\textcolor{premise}P:} Silverware and plate lie on the table. Barbara asked for the plate. {\textcolor{hypothesis}H:} Barbara also asked for the silverware.\\
\midrule
T172 & {\textcolor{premise}P:} \{OBJ1\} and \{OBJ2\} lie on the table. \{NAME\} asked for \{OBJ1\} {\textcolor{hypothesis}H:} \{NAME\} also asked for \{OBJ2\} {\colorbox{pink}{\texttt{neutral}}} & \multirow{2}{*}{33.80} & \multirow{2}{*}{18.40} & \multirow{2}{*}{99.80}\\
Bool & {\textcolor{premise}P:} Pin and pitcher lie on the table. Peter asked for the pitcher. {\textcolor{hypothesis}H:} Peter also asked for the pin.\\
\midrule
T128 & {\textcolor{premise}P:} Some the \{OBJS\} are \{COLOR\} in color. {\textcolor{hypothesis}H:} All of the \{OBJS\} are \{COLOR\} in color. {\colorbox{pink}{\texttt{cont}}} & \multirow{2}{*}{100.00} & \multirow{2}{*}{100.00} & \multirow{2}{*}{100.00}\\
Impl & {\textcolor{premise}P:} Some of the balls are purple in colour. {\textcolor{hypothesis}H:} All of the balls are purple in colour.\\
\midrule
T127 & {\textcolor{premise}P:} Some the \{OBJS\} are \{COLOR\} in color. {\textcolor{hypothesis}H:} All of the \{OBJS\} are \{COLOR\} in color. {\colorbox{pink}{\texttt{neutral}}} & \multirow{2}{*}{0.00} & \multirow{2}{*}{0.00} & \multirow{2}{*}{0.00}\\
Quan & {\textcolor{premise}P:} Some of the cars are green in colour. {\textcolor{hypothesis}H:} All of the cars are green in colour.\\
\arrayrulecolor{black}\bottomrule
\end{tabular} 
}
\caption{We show some interesting Templates, and model accuracies on the templates.}
\label{tab:app:interestingtemplates2}
\end{table*}

\end{document}